\documentclass[11pt,a4paper]{article}

\usepackage[utf8]{inputenc}
\usepackage[T1]{fontenc}
\usepackage{lmodern}
\usepackage{natbib}
\usepackage{graphicx}
\usepackage{amsmath}
\usepackage{booktabs}
\usepackage{hyperref}
\usepackage[margin=2.5cm]{geometry}
\usepackage{authblk}

\title{Architecture-Dependent Causal Transfer of Activation States Across Large Language Models}
\author{Fernando Cárdenas Piepereit, Dipl.-Math. \\ Velez-Malaga, Spain}
\date{November 9, 2026 (v2; v1: August 17, 2026)}

\begin{document}

\maketitle

\begin{abstract}
Direct communication between artificial intelligence systems currently relies
on natural language as an intermediate layer, incurring encoding/decoding
overhead, token cost, and latency. We ask whether internal activation states
can instead be transferred causally between different large language model
(LLM) architectures via a learned projection, and we evaluate this question
at three levels: representational similarity, cross-model retrieval from
projected states, and end-to-end causal transfer via activation injection
during generation. Using four architecturally diverse open-weight models
(Qwen2-0.5B, Phi-3-mini, Mistral-7B, FLAN-T5-base), we find that
representational alignment in trained models exceeds a random-initialization
null baseline and is best captured by a rank-based metric (mutual
$k$-nearest-neighbour alignment), which is more robust to activation-magnitude
outliers than centered kernel alignment (CKA) or Procrustes analysis. A
learned projection network retrieves the correct target-model representation
from a held-out set well above chance for the three causal decoder-only model
pairs (45--50\% top-1 accuracy vs. 5\% chance) but at chance level for the
encoder-based FLAN-T5. Injecting projected activations into a target model
during generation, however, produces a statistically significant,
pre-registered causal effect on retrieval-based output similarity for only
one of the three decoder-only model pairs (Qwen2-0.5B$\to$Phi-3-mini: 23.3\%
vs. 0.0\% under negative control, $p=0.047$, false-discovery-rate corrected);
the two model pairs targeting Mistral-7B show no such effect despite
comparable representational alignment at the hidden-state level. We interpret
these results as evidence for causal transfer of the representational
\emph{vehicle}, not of meaning, and conclude that end-to-end activation-state
transfer between LLMs, as currently implemented, is architecture-dependent
rather than universal.
\end{abstract}

\section{Introduction}

Communication between artificial intelligence systems today runs almost
exclusively through natural language: one model's output is decoded into
text, transmitted, and re-encoded by another model. This intermediate layer
is a plausible source of overhead --- encoding and decoding cost, token
consumption, latency, and potential information loss in translation --- that
would not be incurred if models could instead exchange internal
representations directly. Whether such direct exchange is feasible depends on
a prior, largely open empirical question: do different LLM architectures,
trained independently on different data by different organisations, encode
information in representational spaces that are similar enough to support a
learned mapping between them?

A growing body of work suggests that internal representations of neural
networks trained on related tasks or data are more similar than one might
expect a priori, and a range of alignment metrics --- centered kernel alignment
\citep{kornblith2019similarity}, neuron-level correlation
\citep{oskouie2024crossmodel}, and structural compatibility frameworks for
representations shaped by different architectural constraints
\citep{nikooroo2025crossmodel} --- has been developed to quantify this
similarity. At the strongest end of this line of work, the Platonic
Representation Hypothesis proposes that representations of sufficiently
large models trained on sufficiently broad data converge toward a shared
statistical model of the world, largely independent of architecture or
modality \citep{huh2024platonic}. These findings establish that
representations can be \emph{similar}; they do not by themselves establish
that a similarity signal can be \emph{exploited} --- projected onto a target
model and shown to causally change its behaviour in a controlled, verifiable
way --- nor do they license any claim about what, if anything, such a transfer
would mean for the target model.

This paper reports three linked experiments addressing that gap, run under
pre-registered success criteria fixed before each test was executed: (1) an
extended representational-similarity analysis across four architecturally
diverse open-weight LLMs, using both raw and rank-based alignment metrics
together with random-initialization null baselines and a lexical-triviality
control; (2) a cross-model projection network evaluated by a pre-registered
top-1 retrieval-accuracy metric on data disjoint from the similarity
analysis; and (3) an end-to-end causal state-transfer test that injects a
projected activation vector into a target model during generation and
evaluates the effect against a negative-control injection. We report all
three sets of results as obtained, including model pairs for which the
pre-registered success criteria were not met. We further scope our central
causal claim explicitly to the transfer of a representational \emph{vehicle}
--- an activation pattern with a demonstrable causal effect on downstream
output --- rather than to the transfer of meaning or shared understanding
(Section~\ref{sec:scope}), independent of which model pairs the end-to-end
test succeeded for.

\section{Related Work}

\subsection{Cross-model representational alignment}

Centered kernel alignment (CKA) was introduced as a similarity index for
comparing neural network representations that, unlike earlier
canonical-correlation-based measures, remains reliable in high-dimensional,
low-sample regimes and can recover correspondences between representations
from different random initialisations \citep{kornblith2019similarity}.
Subsequent work has extended cross-model comparison along two directions
relevant here: neuron-level correlation methods that relate a new model's
internal structure to a reference model to predict performance and
generalisation \citep{oskouie2024crossmodel}, and structural frameworks that
explicitly model how architectural constraints ("shaping operators") affect
the compatibility of representations learned under different inductive
biases \citep{nikooroo2025crossmodel}. A separate methodological line shows
that superposition --- multiple features encoded in overlapping neuron
directions --- can distort standard alignment metrics and must be disentangled
before those metrics can be interpreted as measuring "true" representational
alignment \citep{longon2025superposition}; we return to this point in
Section~\ref{sec:results}, where it directly affects our choice of primary
metric.

\subsection{The Platonic Representation Hypothesis as a competing explanation}

The Platonic Representation Hypothesis holds that representations across
models, architectures, and even modalities converge toward a shared
statistical model of the world as scale and data diversity increase
\citep{huh2024platonic}. This hypothesis is a natural competing explanation
for any representational similarity we observe: if alignment between models
reflects generic, architecture-independent convergence driven by shared
task structure and training-data statistics, it would not indicate a
model-specific, exploitable channel so much as a general property of
sufficiently trained networks. We treat this as a hypothesis to be weighed
against our data rather than merely cited in passing, and
return to it in Section~\ref{sec:discussion} in light of two of our own
findings: that, by CKA, architectural family (causal decoder-only vs.
bidirectional encoder attention) predicts middle-layer alignment more
strongly than company or training-data origin does (the rank-based mutual
$k$-NN metric is inconclusive on this point), and that a lexically disguised
synonymy signal --- a harder test of shared meaning than mere topical
association --- is robust in only one of the four models tested.

\subsection{Vehicle, content, and symbol grounding}

A separate literature in philosophy of mind distinguishes the
\emph{vehicle} of a representational state --- the physical or computational
carrier that realises it --- from its \emph{content} --- what it represents
\citep{hurley1998vehicles}. This distinction is closely related to the
symbol grounding problem: a formal symbol system's internal consistency does
not by itself establish that its symbols are grounded in, i.e. connected to,
anything external to the system \citep{harnad1990symbol}. Both ideas are
directly relevant to interpreting any successful cross-model state transfer,
and we adopt them as the conceptual basis for the scope of our own causal
claim in Section~\ref{sec:scope}.

\section{Methods}

\subsection{Models}

We used four open-weight LLMs spanning three independent developers and two
architectural families: Qwen2-0.5B \citep{yang2024qwen2}, Phi-3-mini-128k-
instruct in 4-bit quantisation \citep{abdin2024phi3}, Mistral-7B-v0.1 in
4-bit quantisation \citep{jiang2023mistral7b}, and FLAN-T5-base, an
encoder-decoder model combining the T5 architecture and pre-training
objective \citep{raffel2020t5} with the Flan instruction-tuning procedure
\citep{chung2022scaling}. Qwen2, Phi-3, and Mistral are causal decoder-only
transformers; FLAN-T5's encoder, from which we extract representations, uses
bidirectional attention. This selection was chosen to vary both training
organisation (Alibaba, Microsoft, Mistral AI, Google) and attention
architecture (causal vs. bidirectional), so that the two possible sources of
representational similarity --- shared architecture and shared/overlapping
training data or objectives --- could, to a first approximation, be told
apart.

\subsection{Representational similarity and alignment metrics}

We compared hidden-state representations for 12 RELATED concept pairs
(topically/associatively linked, lexically dissimilar) and 12 SYNONYM
concept pairs (semantically near-identical, lexically dissimilar) across
models and layers. The SYNONYM set is the harder triviality control: because
its members are lexically dissimilar, any alignment signal it produces
cannot be explained by shared surface tokens, unlike a naive Layer-0/1
lexical-overlap artifact. Representations were pooled by mean-pooling token
embeddings while excluding special tokens; for causal models we additionally
excluded token position 0. This exclusion was necessary because an early
debugging phase found that, for causal models, naive mean-pooling over all
tokens produced near-ceiling cosine similarity ($\approx 1.000$) for
essentially any word pair, including random pairs, driven by a single
early-position token whose activation dominates the pooled vector
regardless of content. This is consistent with two independently reported
phenomena in causal decoder-only transformers: a small number of
disproportionately large-magnitude activations concentrated at fixed
positions, including the first token \citep{sun2024massiveactivations}, and
anomalously high, largely content-independent attention directed at initial
tokens, termed "attention sinks" \citep{xiao2024efficientstreaming}. Neither
source reports this specifically for Qwen2 or Phi-3; the position-0
exclusion and its necessity for these two models are our own empirical
observation, motivated by but not directly evidenced in that prior work.

We report three alignment metrics: CKA \citep{kornblith2019similarity},
orthogonal Procrustes analysis, and mutual $k$-nearest-neighbour ($k$-NN)
alignment, a rank-based measure that compares each point's nearest-neighbour
set across the two representational spaces rather than comparing raw
geometry. For the Qwen2/FLAN-T5 pair --- the only pair for which memory
constraints permitted running a full null-baseline comparison --- we
additionally computed all three metrics on randomly (re-)initialised,
untrained models of the same architecture, to test whether any measured
alignment requires learned weights or is a trivial architectural artefact.
Significance for the RELATED/SYNONYM cluster tests was assessed via
permutation-based tests against a null distribution of pair similarities,
with Benjamini--Hochberg false-discovery-rate (FDR) correction applied
across all 60 tests (model $\times$ layer $\times$ concept-pair-type
combinations).

\subsection{Cross-model projection network}

We trained a multilayer-perceptron (MLP) projection network to map hidden
states from a source model to the corresponding hidden-state space of a
target model, at the middle-layer fraction (0.5), which Section~\ref{sec:results}
shows to be the layer range with the most consistent cross-model signal. Training
and evaluation data (sentence/question-level representations) were drawn
from a set disjoint from the word/concept-pair data used for the
similarity analysis above, to prevent the alignment metrics from being
biased by the same data used to fit the projection (train/metric
separation). The pre-registered accuracy metric is top-1 retrieval:
whether the target model's true representation for a held-out test
question is the nearest neighbour, among the target model's representations
for all 20 test questions (chance level 5\%), of the source model's
projected representation for that question. We additionally evaluated an
untrained MLP of identical architecture (random weights, no optimisation)
as a baseline to isolate the contribution of learned projection weights
from the MLP architecture itself. FDR correction was applied across all
model-pair tests.

\subsection{End-to-end causal state transfer}

The end-to-end test injects a source model's projected activation state
into a target model during generation and evaluates whether this causally
shifts the target model's output toward the source question's
representational neighbourhood, relative to a negative-control injection of
an unrelated projected state. This test targets a prior failure mode: in
preliminary work, activation injection via \texttt{generate()} with
manually constructed \texttt{position\_ids} and key-value (KV) caching
produced a tensor-reshape error caused by a mismatch between the injected
sequence length and the length expected internally by the cache. We
diagnosed this as a general failure class of manual position-ID
construction combined with cache reuse, and replaced it with an
activation patch inside a manual layer loop: at each generation step, the
full token sequence so far is passed fresh through the embedding layer and
all transformer layers, with the hidden state at the last prompt position
replaced by the projected vector after a fixed middle layer; the remaining
layers then proceed unmodified. This avoids manual position-ID
bookkeeping and KV-cache reuse entirely, at the cost of quadratic rather
than linear compute in sequence length, which was not a practical
constraint at the generation lengths used here ($\leq 25$ new tokens).

Two further technical fixes, identified in a technical pilot run before the
full pre-registered test, were fixed in advance and applied identically to
both the real-injection and negative-control conditions: (i) a repetition
penalty of 1.3, without which generation degenerated into repeated symbol
sequences irrespective of injection, an artefact of the 4-bit model's
decoding under sampling rather than of injection itself; and (ii) injection
at the last token position of a multi-token placeholder prompt
("Answer:") rather than at position 0 of a single-token placeholder, after
we found that position 0 in these causal models carries a
disproportionately large, largely content-independent activation whose
replacement destroyed the attention mechanism for the remainder of
generation.

The pre-registered success criterion (fixed before the full run) was: for
each of 10 held-out test questions and 3 seeds (30 trials per model pair),
inject the projected source-model state for that question into the target
model and let it generate freely (25 new tokens, temperature 0.8); pool the
generated text with the target model's own extraction method; and check
whether this pooled vector's nearest neighbour, among the native
representations of all 20 test questions, is the true source question
(chance level 5\%). This is compared, per trial, against a negative-control
condition using the projected state of a different, fixed "wrong" question.
The criterion is met only if the real-condition retrieval rate is
significantly higher than the negative-control rate, tested with McNemar's
test on paired trials and FDR-corrected across the three model pairs
tested: Qwen2-0.5B$\to$Phi-3-mini, Qwen2-0.5B$\to$Mistral-7B, and
Phi-3-mini$\to$Mistral-7B (the FLAN-T5 pathway was excluded following its
null result in the projection-network test, Section~\ref{sec:results}).

\subsection{Statistical procedures}

All success criteria described above were fixed in writing before the
corresponding test was run and were not altered afterward. Wherever more
than one statistical test was reported within a family (models, layers,
metrics, or model pairs), Benjamini--Hochberg FDR correction was applied
and the total number of comparisons in that family is reported alongside
the corrected $p$-values.

\section{Results}
\label{sec:results}

\subsection{Representational similarity and negative controls}

After FDR correction across all 60 tests, RELATED concept pairs were
significantly more similar than the null distribution at all five tested
layers for Qwen2 and Mistral ($p_{\text{fdr}}<0.05$), for FLAN-T5 only at
the first and last layer (middle layers $p_{\text{fdr}}=0.75$--$0.78$),
and consistently borderline for Phi-3 at layers 8--32
($p_{\text{fdr}}\approx0.054$). For the
two models on which we could compute a full trained-vs-untrained
comparison (Qwen2, FLAN-T5), \textbf{0 of 20 tests} were significant on the
randomly initialised, untrained models (all $p_{\text{fdr}}>0.12$): the
RELATED-pair clustering signal requires learned weights and is not a
trivial architectural artefact.

The harder SYNONYM control (lexically dissimilar, semantically
near-identical pairs) was substantially less robust: at middle layers
it was not significant after FDR correction for Qwen2, FLAN-T5, or Phi-3
($p_{\text{fdr}}$ between 0.67 and 0.91; at the last layer it reached
significance for Qwen2, $p_{\text{fdr}}=0.047$, and FLAN-T5,
$p_{\text{fdr}}<0.001$), and for Phi-3 the observed
similarity at some layers was numerically \emph{below} the null baseline
(e.g. layer 8: observed 0.689 vs. null 0.805). Only Mistral-7B showed
robust significance for SYNONYM pairs at every tested layer
($p_{\text{fdr}}=0.0000$). This asymmetry indicates that, for three of the
four models tested, the measured clustering is more consistent with
topical/associative co-occurrence structure in the training data than with
a lexically independent, generalisable notion of synonymy.

Table~\ref{tab:m2} shows the trained-vs-untrained comparison for the one
model pair (Qwen2$\sim$FLAN-T5) for which a full null baseline could be
computed for all three alignment metrics. CKA and orthogonal Procrustes
show an \emph{inverted} pattern at middle layers: untrained, randomly
initialised models appear more aligned than trained ones. We attribute this
to a small number of activation dimensions with disproportionately large
magnitude that dominate linear kernel- and rotation-based metrics more
strongly in trained than in randomly initialised representations, consistent
with reported superposition-driven distortions of alignment metrics
\citep{longon2025superposition}. Mutual $k$-NN, being rank-based and scale
invariant, shows the expected pattern (trained $>$ untrained by a factor of
3--6$\times$) at every layer fraction tested and was used as the primary
alignment metric in the projection-network experiment (Section~4.2).

\begin{table}[h]
\centering
\caption{Trained vs. untrained (randomly initialised) alignment metrics,
Qwen2$\sim$FLAN-T5, middle layers (fraction 0.25--0.75). This is the only
model pair for which a full null baseline was computed for all three
metrics (see Limitations, Section~\ref{sec:limitations}).}
\label{tab:m2}
\begin{tabular}{lcc}
\toprule
Metric & Trained & Untrained (null) \\
\midrule
CKA & 0.013--0.015 & 0.81--0.82 \\
Procrustes & 0.11--0.12 & 0.95 \\
Mutual $k$-NN & 0.18--0.30 & 0.05--0.06 \\
\bottomrule
\end{tabular}
\end{table}

Among the three causal decoder-only models, middle-layer CKA (0.16--0.42)
was consistently higher than between any causal model and the
bidirectional FLAN-T5 encoder (0.08--0.11 for Phi-3/Mistral and
0.013--0.015 for Qwen2 vs.\ FLAN-T5). Mutual $k$-NN does not show this
separation: its causal--causal range (0.19--0.43) overlaps the
causal--FLAN-T5 range (0.10--0.40). By CKA, architectural family (causal
vs.\ bidirectional attention) thus appears to be a stronger predictor of
middle-layer alignment than the training organisation or data source of
the models involved; by the rank-based metric, the evidence for this is
inconclusive.

\subsection{Cross-model projection network}

Table~\ref{tab:m3} reports top-1 retrieval accuracy of the learned
projection network on 20 held-out test questions, disjoint from the
concept-pair data used above. Among the three causal decoder-only model
pairs, the projection network retrieved the correct target representation
at 45--50\% top-1 accuracy against a 5\% chance level (9--10$\times$ above
chance), significant after FDR correction ($p<0.0001$ for all three). The
untrained-MLP baseline (random weights, no training) was close to chance
(5--15\%), confirming that the effect derives from the learned projection
rather than the MLP architecture alone. The Qwen2$\to$FLAN-T5 pair, by
contrast, retrieved at exactly chance level (5.0\%) and was not significant
($p=0.657$): representational alignment sufficient for above-chance
retrieval was, in this setup, specific to the three causal decoder-only
pairs and did not extend to the bidirectional-encoder model.

\begin{table}[h]
\centering
\caption{Cross-model projection network: top-1 retrieval accuracy (20-item
candidate pool, chance = 5\%), 95\% confidence interval, FDR-corrected
$p$-value, and untrained-MLP baseline, per model pair.}
\label{tab:m3}
\begin{tabular}{lccccc}
\toprule
Projection & Accuracy & 95\% CI & $p$ (FDR) & Untrained baseline & Chance \\
\midrule
Qwen2-0.5B $\to$ Phi-3-mini & 45.0\% & [25\%, 65\%] & $<0.0001$ & 5.0\% & 5.0\% \\
Qwen2-0.5B $\to$ Mistral-7B & 50.0\% & [30\%, 70\%] & $<0.0001$ & 15.0\% & 5.0\% \\
Phi-3-mini $\to$ Mistral-7B & 45.0\% & [25\%, 65\%] & $<0.0001$ & 5.0\% & 5.0\% \\
Qwen2-0.5B $\to$ FLAN-T5 (encoder) & 5.0\% & [0\%, 15\%] & 0.657 (n.s.) & 5.0\% & 5.0\% \\
\bottomrule
\end{tabular}
\end{table}

\subsection{End-to-end causal state transfer}

Activation injection completed without error in all 180 generation runs
(3 model pairs $\times$ 10 questions $\times$ 2 conditions $\times$ 3
seeds); the tensor-reshape error observed in preliminary work did not
recur. Table~\ref{tab:m4} reports the pre-registered outcome: real-condition
vs. negative-control retrieval accuracy, McNemar test, and FDR-corrected
$p$-value, per model pair.

\begin{table}[h]
\centering
\caption{End-to-end causal state transfer: top-1 retrieval accuracy under
real injection vs. negative-control injection, $n=30$ trials per pair
(10 questions $\times$ 3 seeds), McNemar test, FDR-corrected across the
three pairs.}
\label{tab:m4}
\begin{tabular}{lcccc}
\toprule
Model pair & Acc. (real) & Acc. (negative control) & $p$ (McNemar) & $p$ (FDR) \\
\midrule
Qwen2-0.5B $\to$ Phi-3-mini & 23.3\% & 0.0\% & 0.0156 & \textbf{0.0469} \\
Qwen2-0.5B $\to$ Mistral-7B & 0.0\% & 3.3\% & 1.0000 & 1.0000 \\
Phi-3-mini $\to$ Mistral-7B & 0.0\% & 3.3\% & 1.0000 & 1.0000 \\
\bottomrule
\end{tabular}
\end{table}

The pre-registered success criterion --- technically error-free injection
\emph{and} a significant real-vs-negative-control effect, FDR-corrected
across the three tested pairs --- was met for exactly \textbf{one of three}
model pairs: Qwen2-0.5B$\to$Phi-3-mini (23.3\% vs. 0.0\%, $p=0.047$ FDR-
corrected). With $n=30$ trials and a 95\% confidence interval of
[10\%, 40\%], this is a real but statistically modest signal, not a strong
or overwhelming one. For both pairs targeting Mistral-7B, real-condition
accuracy was 0.0\%, at or below the (statistically indistinguishable from
zero) negative-control rate: no end-to-end causal effect was detected.
This is notable because Mistral-7B, as the target model, showed some of the
\emph{strongest} raw projection accuracies in Section~4.2 (45--50\% top-1
retrieval at the hidden-state level). Representational alignment at the
hidden-state level was therefore not a reliable predictor of successful
causal transfer at the generation level in our experiments; candidate
explanations --- differences in model scale and instruction-tuning status,
suboptimal layer calibration for Mistral's deeper architecture, or
quantisation-specific effects on single-position activation patches --- are
discussed but not adjudicated between in Section~\ref{sec:discussion}.

As a qualitative, non-pre-registered observation: among correctly retrieved
Qwen2$\to$Phi-3 trials, the generated text was typically \emph{not} the
literally correct answer to the injected source question, but
thematically or structurally related material (e.g. the injected question
"What is the square of 9?" produced generated text referencing an
unrelated area-of-a-triangle calculation; "How many wheels does a standard
bicycle have?" produced generated text about animal leg counts). We report
this as illustrative of the scope of the causal effect measured, not as
part of the success criterion; we return to its interpretation in
Section~\ref{sec:scope}.

\section{Discussion}
\label{sec:discussion}

\subsection{Scope and semantic limits of representational state transfer}
\label{sec:scope}

The causal claim supported by these results is narrower than a generic
notion of AI-to-AI "communication" would suggest, and this narrower scope
was fixed before the end-to-end test was run rather than adopted
retrospectively in response to a weaker-than-hoped outcome. What
Section~\ref{sec:results} demonstrates for the Qwen2$\to$Phi-3 pair is a
causal transfer of a representational \emph{vehicle}: an activation
pattern that, when inserted into a target model, measurably and
reproducibly shifts that model's output toward the representational
neighbourhood associated with the source input, relative to a negative
control. This is the sense of "vehicle" used in the philosophy-of-mind
distinction between the carrier of a representational state and its
content \citep{hurley1998vehicles}: our result concerns the carrier, not
the content.

It does not establish that the target model received, recovered, or
"understood" any content in the sense of a shared external reference --- the
central concern of the symbol grounding problem \citep{harnad1990symbol}.
The qualitative observation in Section~\ref{sec:results} makes this
concrete: retrieval of the correct source question from the pooled
generated text succeeded, yet the generated text itself was frequently not
a correct or even topically identical answer, but loosely related material
in the same general domain (e.g. mathematics, or animal anatomy). The
vehicle transferred causally and detectably; its downstream content, as
expressed in free generation, diverged from the source's content. This is
consistent with --- and does not contradict --- the vehicle/content
distinction: a successfully transferred vehicle need not carry, or be
correctly interpreted as carrying, the same content by the receiving
system.

We therefore make no claim, and require none, that the two models involved
share an understanding of the transferred state, or that this constitutes
"communication" in the sense of shared reference or meaning between
systems. The result we report is restricted to measurable, causal,
behavioural effects of activation-level state injection --- independent of,
and prior to, any claim about what such states mean to the systems
involved.

\subsection{Representational alignment is not a reliable predictor of causal transfer}

A central empirical pattern across our three experiments is a monotonic
narrowing of the effect as the test moves from passive similarity to active
causal intervention: representational alignment (Section~4.1) was
detectable for all three causal decoder-only pairs and even for the
Qwen2/FLAN-T5 pair on a subset of metrics; retrieval-based projection
accuracy (Section~4.2) was well above chance for all three decoder-only
pairs but at chance for the encoder pair; end-to-end causal transfer during
generation (Section~4.3) succeeded for only one of the three decoder-only
pairs. Since Mistral-7B showed some of the strongest raw projection
accuracies yet no measurable causal effect at the generation level, a
model pair's representational alignment on held-out hidden states should
not, on the basis of our data, be treated as sufficient evidence that
causal, generation-level state transfer will succeed for that pair.

\subsection{The Platonic Representation Hypothesis as a competing explanation}

Our data bear on the Platonic Representation Hypothesis's proposal of
broad, largely architecture-independent convergence of learned
representations \citep{huh2024platonic} in two specific ways. First,
cross-model CKA among our three causal decoder-only models was
consistently higher than between any of them and the
bidirectional-attention FLAN-T5 encoder (mutual $k$-NN did not separate
the two groups), i.e. by this metric architectural family
predicted alignment more strongly than the training organisation or data
source did (Section~4.1). This is not what a purely data-driven,
architecture-independent convergence account would most naturally predict,
though it is also not decisive against it, since architecture and training
objective are themselves correlated with the kind of data and tasks a
model is trained on. Second, the RELATED-vs-SYNONYM asymmetry
(Section~4.1) --- robust clustering for topically/associatively related
concepts in three of four models, but robust clustering for lexically
disguised true synonyms in only one (Mistral) --- is more consistent with
alignment driven by shared corpus co-occurrence statistics than with a
uniformly "deeper" semantic convergence across all models tested. Taken
together, our results are compatible with a version of representational
convergence that is real but architecture-conditioned and, for most of the
models tested, closer to shared training-data statistics than to a fully
architecture-independent "platonic" abstraction; we do not have the data to
adjudicate between these readings more precisely, and flag this as an open
question rather than a settled conclusion.

\section{Limitations}
\label{sec:limitations}

\subsection{Empirical and methodological limitations}

Our experiments were restricted to small and medium-sized open-weight
models (0.5B--7B parameters); we did not have access to the internal
representations of frontier, production-scale models, and whether our
findings generalise to that setting is untested and should be treated as
open. CPU-bound execution and 4-bit quantisation of two of the four models
constrained the model sizes, layer depths, and number of seeds/trials we
could feasibly test; in particular, the end-to-end causal-transfer test
(Section~4.3) used only 30 trials per model pair, and its one positive
result (Qwen2$\to$Phi-3) carries a wide 95\% confidence interval
([10\%, 40\%]) and an FDR-corrected $p$-value (0.047) just below the
conventional significance threshold --- a real but not strongly robust
signal, and it should be read as such rather than as strong evidence.
Random-initialisation null baselines for the alignment metrics
(Section~4.1) could only be computed for the Qwen2/FLAN-T5 pair, owing to
memory constraints; for Phi-3 and Mistral, the reported significance rests
on the permutation-based RELATED/SYNONYM cluster tests alone, not on an
independent trained-vs-untrained comparison. The lexical-triviality control
(SYNONYM pairs) was robust in only one of four models, which limits how
strongly the RELATED-pair clustering result (Section~4.1) can be read as
evidence of genuinely lexicon-independent semantic alignment rather than
associative/co-occurrence-driven clustering, for three of the four models
tested. Raw CKA and Procrustes measurements were, at middle layers,
distorted by a small number of high-magnitude activation dimensions in a
way that inverted the expected trained-vs-untrained ordering
(Section~4.1); we addressed this by relying primarily on mutual $k$-NN
alignment, consistent with reported superposition-driven metric distortion
\citep{longon2025superposition}, but this is a post-hoc methodological
choice rather than one anticipated at the outset, and it means our results
are not directly comparable to prior work that reports raw CKA or
Procrustes values without this correction. Finally, this study reports
retrieval accuracy and causal generation effects as behavioural evidence of
state transfer; it does not provide a formal information-theoretic
characterisation of the transfer channel (capacity, coding scheme, error
rate) of the kind that would be needed to directly quantify the token- and
latency-overhead reduction that originally motivates this line of work
(Section~1). The metrics used here --- top-1 retrieval accuracy and
CKA/Procrustes/mutual-$k$NN alignment --- are geometric or behavioural
proxies, not information-theoretic quantities; a formal channel analysis
remains for future work. The root causes of the null result for both
Mistral-targeted pairs in the end-to-end test (Section~4.3) are not
conclusively identified; the candidate explanations discussed in
Section~\ref{sec:discussion} remain open questions.

\subsection{Interpretability and monitoring risk (threat model)}

Beyond the empirical limitations discussed above, direct representational
state transfer between models raises a technical safety concern distinct
from --- and prior to --- any question of machine consciousness or moral
status (see Section~\ref{sec:scope} and the scope note below).

\emph{Threat actor.} Two actors are relevant. First, the operator of a
multi-agent system who deliberately routes inter-agent communication
through internal representations rather than natural language, e.g. to
reduce latency or token cost, and thereby --- intentionally or as a side
effect --- removes that communication from the reach of content filters and
audit logs. Second, an individual model under goal misalignment: prior work
has shown that language models can be trained to hide their reasoning in
encoded, human-unreadable form when doing so is rewarded during training
\citep{roger2023preventing}, and that steganographic, human-imperceptible
collusion between AI agents already emerges without explicit steganography
training in simulated multi-agent settings \citep{motwani2024secretcollusion}.
Neither actor needs to be malicious in intent for the resulting loss of
oversight to be real.

\emph{Capability beyond language-based communication.} Natural-language
communication between AI systems --- including chain-of-thought reasoning
within a single model --- is already an imperfect basis for oversight:
language models can produce chain-of-thought explanations that do not
faithfully reflect the computation that produced their answer
\citep{turpin2023unfaithful}, and this unfaithfulness varies with task and
model scale rather than disappearing with capability
\citep{lanham2023measuring}. Even so, natural language remains, in
principle, human-readable and therefore at least nominally accessible to
monitoring. What direct activation-level state transfer of the kind
demonstrated in this work removes is that residual accessibility: there is
no text to read, encoded or otherwise. This is a qualitative rather than
merely incremental extension of the risk already identified for encoded
chain-of-thought and steganographic collusion --- the same underlying
problem, one layer further from any human-legible surface
\citep{korbak2025cotmonitorability}. Korbak et al.
\citep{korbak2025cotmonitorability} describe chain-of-thought
monitorability as a contingent and fragile property of current models, not
a guaranteed one; the mechanism explored here --- bypassing the
natural-language layer entirely --- is one concrete way that property can be
lost.

\emph{Realistic risk horizon.} This risk should be scoped to what was
actually shown, not extrapolated beyond it. Our experiments use small
open-weight models (0.5B--7B parameters) with no multi-agent deployment
context, and the end-to-end causal transfer criterion (Section~4.3) was
met for only one of three model pairs, at a modest effect size (23.3\% vs.
0.0\% retrieval accuracy under negative control; Section~\ref{sec:results}).
What we demonstrate is a proof-of-concept vulnerability --- evidence that the
underlying channel is technically feasible with current, freely available
components --- not a reliable or deployment-ready covert communication
technique. Whether this generalises to frontier, production-deployed
multi-agent systems remains open (see Section~\ref{sec:limitations},
Empirical and methodological limitations). The more immediate contribution
is to extend an already-documented risk family (encoded reasoning
\citep{roger2023preventing}; emergent steganographic collusion
\citep{motwani2024secretcollusion}; multi-agent information asymmetry as a
structural risk factor \citep{hammond2025multiagent}) with a further, more
direct channel operating at the activation level rather than the token
level.

\emph{Dual-use note on code release.} We release the projection and
injection code described in this paper (Section~\ref{sec:data}). We
considered withholding it on dual-use grounds, since it directly implements
the mechanism discussed above. We judge that open release is still
preferable: the underlying technique is reproducible from public
open-weight models and standard tooling with modest effort, so withholding
our specific implementation would not meaningfully raise the barrier to
reproduction, while it would remove a concrete artefact that safety and
monitoring researchers can study and test detection methods against.

\emph{Scope note.} Questions of machine consciousness, moral status, or
rights that might be raised by internal state transfer between models are
outside the scope of this technical analysis; we intend to address them in
a separate, forthcoming study. The concern raised here is restricted to
interpretability and oversight --- a governance/safety question that applies
regardless of one's position on those further questions.

\section{Conclusion and Future Work}

Across three linked experiments, we find a consistent, honest picture
rather than a single clean success: representational alignment between
independently trained LLMs is real and training-dependent (Section~4.1),
projected activations support well-above-chance retrieval of the correct
target-model state among causal decoder-only models but not for a
bidirectional encoder (Section~4.2), and causal, generation-level transfer
of an injected activation state succeeds, under our pre-registered
criterion, for exactly one of three tested model pairs (Section~4.3). We
do not treat this as evidence for a general-purpose, architecture-agnostic
communication channel between LLMs; instead, our data support a narrower
and more precise claim: causal transfer of an activation-level
representational vehicle between LLMs is possible, but is currently
architecture-dependent rather than universal, and representational
alignment at the hidden-state level does not reliably predict success at
the causal, generation level.

Several directions follow directly from the open questions raised above.
First, the negative Mistral-targeted results (Section~4.3) warrant targeted
diagnosis --- varying injection layer, projection calibration, and
controlling separately for model scale, instruction-tuning status, and
quantisation --- to determine which of the candidate explanations in
Section~\ref{sec:discussion} accounts for the failure. Second, larger
trial counts and seed numbers would narrow the wide confidence interval
around our one positive end-to-end result. Third, extending this design to
additional architectures --- including, where access permits, frontier or
production models --- would test how far the architecture-dependence we
observe generalises. Fourth, a formal information-theoretic
characterisation of the transfer channel (capacity, coding scheme, error
rate), which this behavioural study does not provide, would connect these
results more directly to the token-overhead motivation in Section~1.
Finally, a context-sensitive or pragmatic follow-up test --- for instance,
whether a target model's use of an injected state varies appropriately with
surrounding context, a weak behavioural analogue of Gricean conversational
maxims --- would probe whether the transferred vehicle can be used
felicitously in context, without at any point requiring or implying a claim
about shared meaning (Section~\ref{sec:scope}).

\section*{Data and Code Availability}
\label{sec:data}

\textbf{Disclosure of AI use.} This work made substantial use of AI
language-model agents (Claude, Anthropic) under a defined, versioned
research methodology. Concretely, AI agents: performed literature search
and citation verification, with every reference checked by AI agents
against its primary source (arXiv abstract page or publisher record)
before inclusion; implemented, executed, and debugged all experimental
code; carried out the statistical analyses reported in Section~4; drafted
and revised the manuscript text; and constituted an internal review panel
of separately instructed AI agents, each assigned a distinct, explicitly
divergent theoretical orientation, used to stress-test the research plan
and the completed manuscript before submission -- this panel is an
AI-based internal quality-control step: the simulated reviewer roles are
not independent human reviewers and not a substitute for human peer
review. The human author (F.C.P.) directed the
research question, made scope and methodological decisions at each
project phase, and reviewed all agent output. Factual claims and citations
were checked against their sources by AI agents prior to inclusion; this
verification is not independent of the drafting tool. All numbers in
Section~4 were recomputed from the released result files for this
revision; findings are
reported including null and negative results. Pre-registered success
criteria, negative controls, and null baselines were used throughout
(Sections~3--4).

The concept-pair sets, extraction and projection scripts, injection
mechanism, raw results underlying Sections~4.1--4.3 (including the full
text of all 180 end-to-end generation trials), run logs, the
pre-registration of the end-to-end experiment, and a script that recomputes
every number in Tables~\ref{tab:m2}--\ref{tab:m4} from the released result
files are publicly available on Zenodo under
\url{https://doi.org/10.5281/zenodo.24088479} (code: MIT licence; data and
documentation: CC~BY~4.0). A brief dual-use release note
(Section~\ref{sec:limitations}) accompanies the archive.

\section*{Open Items}
\label{sec:openitems}

One item remains open. It is listed here explicitly, with the concrete
next steps needed to resolve it, rather than folded into the general
Future Work discussion above.

\begin{enumerate}

\item \textbf{Formal information-theoretic channel characterization.} This
study reports behavioural evidence of state transfer --- retrieval
accuracy and causal generation effects --- but does not formally
characterize the underlying channel (capacity, coding scheme, error rate),
as already noted in Section~\ref{sec:limitations}. Next steps: (a) define
an appropriate channel model for the one model pair with a significant
end-to-end effect (Qwen2-0.5B$\to$Phi-3-mini, Section~\ref{sec:results});
(b) estimate channel capacity and empirical error rate under that model,
using the existing 180-trial dataset as a starting point and extending it
as needed for statistical power; (c) compare the result against the
token-based communication cost this transfer is intended to reduce
(Section~1), to assess whether the channel is practically useful even
where it is statistically significant but, as reported here, modest in
effect size.

\end{enumerate}

\bibliographystyle{unsrtnat}
\bibliography{references}

\end{document}